\documentclass[11pt,a4paper]{article}
\usepackage[hyperref]{acl2019}
\usepackage{times}
\usepackage{latexsym}
\usepackage{booktabs}
\usepackage{graphicx}
\usepackage{amsmath}
\usepackage{url}
\usepackage{threeparttable}
\usepackage{comment,color,soul}
\usepackage{subcaption}

\aclfinalcopy 
\def\aclpaperid{2212} 
\renewcommand{\footnotesize}{\scriptsize}

\title{Explain Yourself! \\Leveraging Language Models for Commonsense Reasoning}

\author{Nazneen Fatema Rajani \quad Bryan McCann \quad Caiming Xiong \quad Richard Socher \\
  Salesforce Research \\
  Palo Alto, CA, 94301\\
  \texttt{\{nazneen.rajani,bmccann,cxiong,rsocher\}@salesforce.com}
  }

\date{Paper ID 2212}

\begin{document}
\maketitle
\begin{abstract}
Deep learning models perform poorly on tasks that require commonsense reasoning, 
which often necessitates some form of world-knowledge or reasoning over information not immediately present in the input.
We collect human explanations for commonsense reasoning in the form of natural language sequences and highlighted annotations in  
a new dataset called Common Sense Explanations (CoS-E).
We use CoS-E to train language models to automatically generate explanations that can be used during training and inference in a novel Commonsense Auto-Generated Explanation (CAGE) framework. 
CAGE improves the state-of-the-art by $10\%$ on the challenging CommonsenseQA task. 
We further study commonsense reasoning in DNNs using both human and auto-generated explanations including transfer to out-of-domain tasks.
Empirical results indicate that we can effectively leverage language models for commonsense reasoning.
\end{abstract}

\section{Introduction} 
\vspace{-0.2cm}
\label{introduction}
Commonsense reasoning is a challenging task for modern machine learning methods \citep{zhong2018improving,talmor2018commonsenseqa}. 
Explanations are a way to verbalize the reasoning that the models learn during training.
Common sense Question Answering (CQA) is a multiple-choice question answering dataset proposed for developing natural language processing (NLP) models with commons-sense reasoning capabilities \citep{talmor2018commonsenseqa}. 
Although these efforts have led to progress, it is still unclear how these models perform reasoning and to what extent that reasoning is based on world knowledge. 
We collect human explanations for commonsense reasoning built on top of CQA and introduce them as Common Sense Explanations (CoS-E)\footnote{\url{https://github.com/nazneenrajani/CoS-E}}. 
CoS-E contains human explanations in the form of both open-ended natural language explanations as well as highlighted span annotations that represent words selected by humans as important for predicting the right answer (see Table~\ref{cose}).

\begin{table}[t]
\centering
\small
\begin{tabular}{ll}
\toprule
Question: &While eating a \hl{hamburger with friends},\\
&what are people trying to do?\\ 
Choices: &\textbf{have fun}, tasty, or indigestion \\
CoS-E: & Usually a hamburger with friends indicates \\
&a good time.\\
\midrule
Question: &\hl{After getting drunk people} couldn't \\
&understand him,it was because of his what?\\ 
Choices: &lower standards,\textbf{slurred speech}, \\
&or falling down\\
CoS-E: & People who are drunk have difficulty speaking.\\
\midrule
Question: &People do what during their \hl{time off} \\
&\hl{from work}?\\
Choices: &\textbf{take trips}, brow shorter, or become hysterical \\
CoS-E: & People usually do something relaxing, such as  \\
&taking trips,when they don't need to work.\\
\bottomrule
\end{tabular}
\caption{Examples from our CoS-E dataset.}
\vspace{-0.5cm}
\label{cose}
\end{table}
 
\citet{talmor2018commonsenseqa} show that using Google search to extract context from top \(100\) result snippets for each of the question and answer choices does not help much in improving the accuracy on CQA trained using even the state-of-the-art reading comprehension model BiDAF++ \citep{Seo2017BidirectionalAF} augmented with a self-attention layer and ELMo representations \citep{peters2018deep}. 

 \begin{figure*}[t!]
    \centering
    \begin{subfigure}[t]{0.49\textwidth}
        \centering
        \includegraphics[width=0.49\textwidth,]{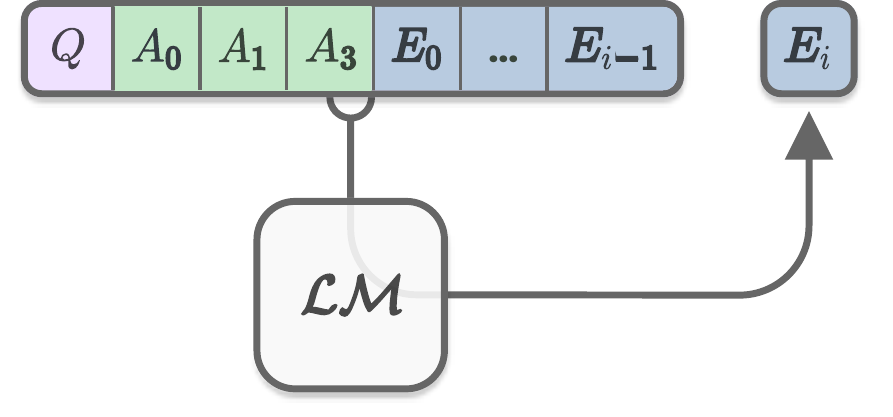}
        \caption{One time-step of training a CAGE language model to generate explanations from CoS-E. It is conditioned on the question tokens $\mathcal{Q}$ concatenated with the answer choice tokens $A_1,A_2,A_3$ and previously generated tokens $E_1,\ldots,E_{i-1}$. It is trained to generate token $E_i$.}
    \end{subfigure}%
    ~ 
    \begin{subfigure}[t]{0.49\textwidth}
        \centering
        \includegraphics[width=0.49\textwidth]{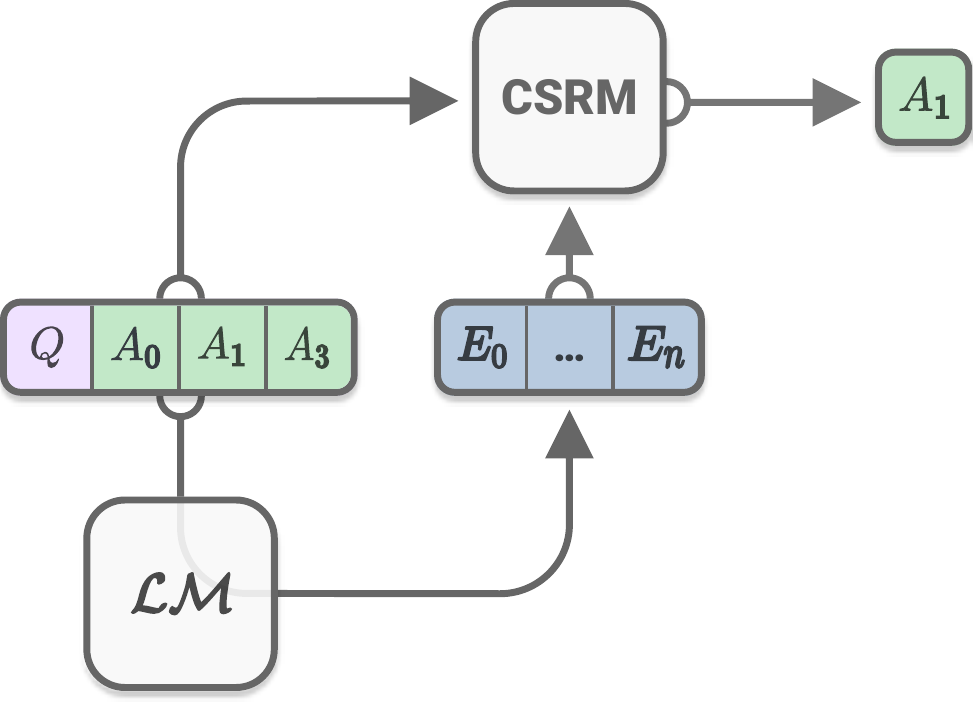}
        \caption{A trained CAGE language model is used to generate explanations for a downstream commonsense reasoning model (CSRM), which itself predicts one of the answer choices.}
    \end{subfigure}
    \caption{An overview of CAGE trained on CoS-E and CQA.}\label{model}
    \vspace{-0.5cm}
\end{figure*}
In contrast, we leverage a pretrained language model to generate explanations that are useful for commonsense reasoning. 
We propose Commonsense Auto-Generated Explanations (CAGE) as a framework for generating explanations for CQA.
We break down the task of commonsense reasoning into two phases.
In the first phase, we provide a CQA example alongside the corresponding CoS-E explanation to a language model.
The language model conditions on the question and answer choices from the example and is trained to generate the CoS-E explanation.

In the second phase,
we use the language model to generate explanations for each example in the training and validation sets of CQA. 
These CAGE explanations are provided to a second commonsense reasoning model by concatenating it to the end of the original question, answer choices, and output of the language model. The two-phase CAGE framework obtains state-of-the-art results outperforming the best reported baseline by $10\%$ and also produces explanations to justify its predictions. Figure~\ref{model} shows an overview of our proposed approach. 

In summary, we introduce a new Common Sense Explanations (CoS-E) dataset to study neural commonsense reasoning and provide a new method, CAGE for automatically generating explanations that achieve a state-of-the-art accuracy of approximately $65\%$ on CQA v1.0. We demonstrate explanation transfer on two out-of-domain datasets. Note that before our final submission, the organizers released a more challenging v1.11 of CQA with \(5\) answer choices instead of \(3\) and so we also included the new version in our results and discussions.

\section{Background and Related Work}
\vspace{-0.2cm}

\paragraph{Commonsense reasoning} 
Datasets that require models to learn to predict relations between situations or events in natural language have been introduced in the recent past.
The Story Cloze (also referred to as ROC Stories) involves predicting the correct story ending from a set of plausible endings \citep{mostafazadeh-EtAl:2016:N16-1} while the Situations with Adversarial Generations (SWAG) involves predicting the next scene based on an initial event \citep{zellers2018swag}. 
Language Modeling based techniques such as the GPT and BERT models get human-level performance on these datasets \citep{radford2018improving,devlin2018bert}. 
They have been less successful on tasks that require clear understanding of how pronouns resolve between sentences and how that interacts with world knowledge.
For example, the Winograd Schemas \citep{winograd1972understanding} and challenges derived from that format \citep{levesque2012winograd,mccann2018natural,wang2018glue} have proven difficult for even the most modern machine learning methods \citep{trinh2018simple} to achieve near-human performance, but the emphasis on pronoun resolution in those challenges leaves room for exploration of other kinds of commonsense reasoning.
CQA is a new dataset that consists of $9500$ questions with one correct answer and two distractor answers \citep{talmor2018commonsenseqa}. 
The authors claim that because all the answer choices are drawn from the same source concept, the dataset requires models to actually infer from the question rather than take advantage of distributional biases. We, however, observed that the current state of this dataset has gender disparity with higher proportion of feminine pronouns used in negative context.

The authors show that the state-of-the-art language models perform very poorly compared to human participants on their dataset. Although, CQA introduces a benchmark for evaluating commonsense reasoning capabilities of models, it is still unclear how and to what extent do models actually do common-sense reasoning. CoS-E builds on top of their benchmark, on the other hand, provides data in the form of explanations that can be used to study and analyze as well as evaluate a model's reasoning capabilities.

\paragraph{Natural language explanations} \citet{lei2016rationalizing} proposed an approach for rationale generation for sentiment analysis by highlighting complete phrases in the input text that by itself is sufficient to predict the desired output. Human-generated natural language explanations for classification data have been used in the past to train a semantic parser that in turn generates more noisy labeled data which can used to train a classifier \citep{hancock2018training}. 
\citet{camburu2018snli} generate explanations and predictions for the natural language inference problem~\citep{camburu2018snli}. 
However, the authors report that interpretability comes at the cost of loss in performance on the popular Stanford Natural Language Inference \citep{bowman2015large} dataset. 
We find that, unlike for e-SNLI, explanations for CQA lead to improved performance in what \citet{camburu2018snli} would call the explain-predict setting. In the multi-modal setting, \citet{rajani2018stacking} showed that visual explanations can be leveraged to improve performance of VQA \citep{vqa} and that an ensemble explanation is significantly better than individual explanations using both automated and human evaluations \citep{rajani:vigil17}.

\paragraph{Knowledge Transfer in NLP}
Natural language processing has often relied on the transfer of world-knowledge through pretrained word vectors like Word2Vec~\citep{mikolov2013efficient} and GloVe~\citep{pennington2014glove}.
Contextualized word vectors~\citep{mccann2017learned,peters2018deep} refined these representations for particular inputs by using different forms of general encoding. 
Language models trained from scratch on large amounts of data have made groundbreaking success in this direction by carefully fine-tuning for specific tasks~\citep{dai2015semi,radford2018improving,howard2018universal,devlin2018bert}. 
These models have the advantage that only a few parameters need to be learned from scratch and thus 
perform surprisingly well even on small amounts of supervised data. 
Fine-tuned language models do not however 
work as well for directly predicting answers for CQA~\citep{talmor2018commonsenseqa}. 
In our work,
we show how these fine-tuned language models are more effective when leveraged to generate explanations and empirically prove that they also linguistically capture common sense.
\section{Common Sense Explanations (CoS-E)}
\vspace{-0.2cm}
\label{sec:cose}

We used Amazon Mechanical Turk (MTurk) to collect explanations for our Common Sense Explanations (CoS-E) dataset. 
The CQA dataset consists of two splits -- the {\it question token split} and the {\it random split}.
Our CoS-E dataset and all our experiments use the more difficult {\it random split}, which is the main evaluation split according to \citet{talmor2018commonsenseqa}. We also release CoS-E for CQA v1.11.

Human participants are given the question and answer choices along with the ground-truth answer choice. 
Turkers are prompted with the following question:
``Why is the predicted output the most appropriate answer?" 
Annotators were instructed to highlight relevant
words in the question that justifies the ground-truth answer choice
and to provide
a brief open-ended explanation based on the
highlighted justification could serve as the commonsense reasoning behind the question.
We collected these explanations for the CQA train-random-split and dev-random-split, 
which have a size of $7610$ and $950$ for v1.0 and $9741$ and $1221$ for v1.11 respectively.
Table~\ref{cose} shows a random sample of examples from our CoS-E dataset with both free-form explanations and highlighted text.
From here on, we refer to the highlighted words as \textit{CoS-E-selected} and the free-form explanation as \textit{CoS-E-open-ended}.
 
In MTurk, it is difficult to control the quality of open-ended annotations. 
So, we do some in-browser checks to avoid obviously bad explanations. 
Annotators cannot move forward if they do not highlight any relevant words in the question or if the length of explanations is less than $4$ words.
We also check that the explanation is not a sub-string of the question or the answer choices without any other extra words.
We collect these explanations from only one annotator per example,
so we also perform some post-collection checks to catch examples that are not caught by our previous filters. 
We filter out explanations that could be classified as a template.
For example, explanations of the form ``$<$answer$>$ is the only option that is $[$correct$\vert$obvious$]$" are deleted and then re-annotated.
 \begin{figure}
 \centering
 \includegraphics[width=0.5\textwidth]{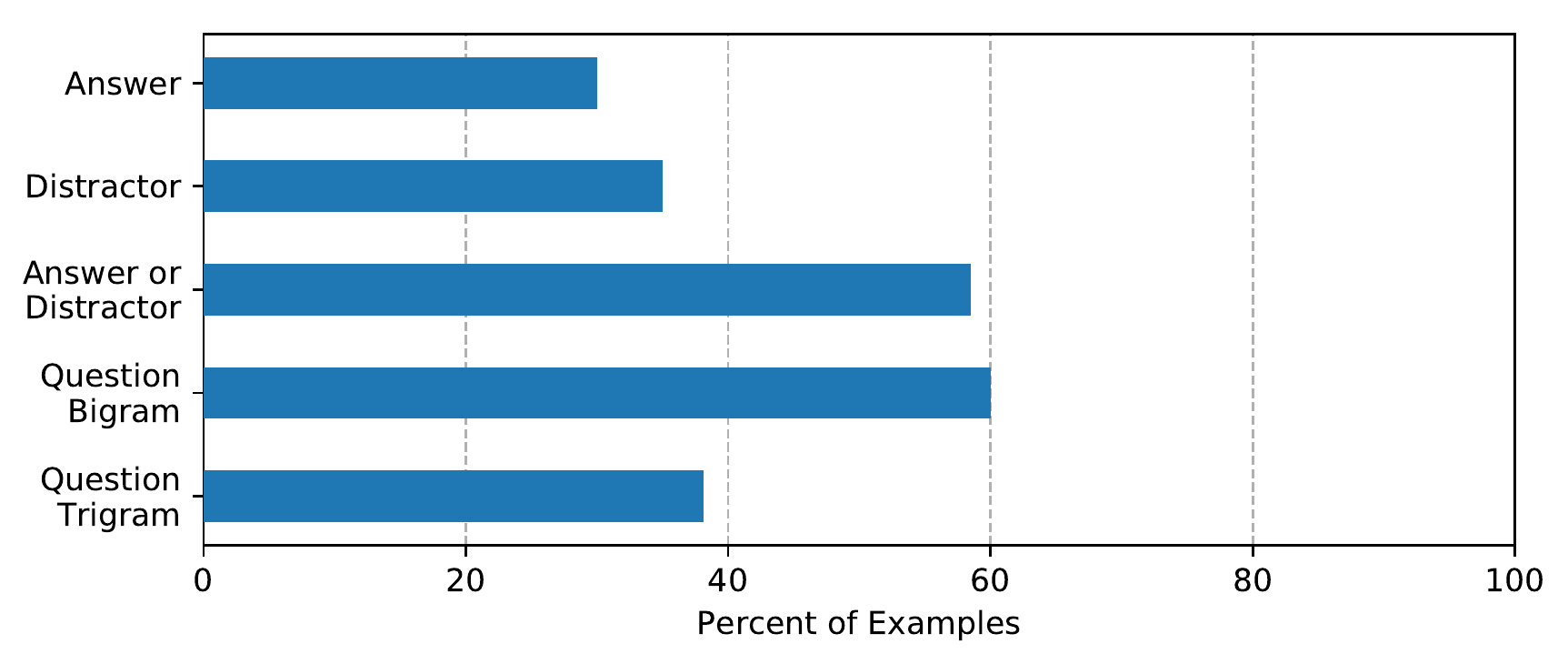}
\caption{Analysis of the CoS-E v1.0 dataset. Percent of the dataset that contains the answer, a distractor, either, at least one bigram from the question, and at least one trigram from the question.} \label{expl-analysis}
\vspace{-0.5cm}
 \end{figure}


Figure~\ref{expl-analysis} shows the distribution of explanations collected in the CoS-E v1.0 dataset. 
$58\%$ of explanations from CoS-E contain the ground truth, but the effectiveness of CoS-E is not constrained only to those examples.
Our model obtains state-of-the-art results by using CoS-E only during training.
Empirical results show that even when using only those explanations that do {\it not} have {\it any} word overlap with {\it any} of the answer choices,
performance exceeds that of baselines that do not use CoS-E at all.
We also observed that a significant proportion of the
distractor choices are also present in the CoS-E
dataset and on further analysis we found that for
those examples, annotators resorted to explaining
by eliminating the wrong choices. This indicates
that it is difficult even for humans to reason about
many of the examples in CQA. Because CoS-E uses crowd-sourcing, it also adds diversity
of perspective and in particular diverse reasoning
on world knowledge to the CQA
dataset.
Even though many explanations remain noisy after quality-control checks, 
we find that they are of sufficient quality to train a language model that generates commonsense reasoning. 
We refer to Section~\ref{experiments} for more details on empirical results and ablation analysis on CoS-E.

\section{Algorithm}
\vspace{-0.2cm}
We present Commonsense Auto-Generated Explanations (CAGE) and apply it to the CQA task. 
CAGE are generated by a language model and are used aas supplementary inputs to a classification model. 
Each example in CQA consists of a question, $q$, three answer choices, $c0, c1, c2$, and a labeled answer $a$. Our CoS-E dataset adds a human explanation $e_{h}$ for why $a$ is the most appropriate choice. 
The output of CAGE is a language model generated explanation $e$ that is trained to be close to $e_{h}$.

\subsection{Commonsense Auto-Generated Explanations (CAGE)}\label{sec:generating}
\vspace{-0.2cm}
In order to supply CAGE to a classification model, we fine-tune a language model (LM) to generate explanations from our CoS-E dataset.
Our LM is the large, pre-trained OpenAI GPT \citep{radford2018improving} which is a multi-layer, transformer \citep{vaswani2017attention} decoder. 
GPT is fine-tuned on the combination of CQA and CoS-E datasets, as shown in the left half of Figure~\ref{model}.
We explore explanation generation in two settings -- 1) explain-and-then-predict (reasoning) (Figure~\ref{model}) and 2) predict-and-then-explain (rationalization).

\paragraph{Reasoning}
This is our main approach and in this the LM is fine-tuned conditioned on the question, answer choices and the human generated explanation and not the actual predicted label. So, the input context during training is defined as follows:
\begin{center}
$C_{RE} = $ ``$q$, $c0$, $c1$, or $c2$? commonsense says "
\end{center}
\noindent
The model is trained to generate explanations $e$ according to a conditional language modeling objective.  The objective is to maximize:
\[
\sum_i \log P(e_i | e_{i-k}, \ldots, e_{i-1}, C_{RE}; \Theta)
\]
\noindent
where $k$ is the size of the context window (in our case $k$ is always greater than the length of $e$ so that the entire explanation is within the context). 
The conditional probability $P$ is modeled by a neural network with parameters $\Theta$ conditioned on $C_{RE}$ and previous explanation tokens.
We call this kind of explanation {\it reasoning} because they can be automatically generated during inference to provide additional context for commonsense question answering. In Section~\ref{experiments}, we show that this approach outperforms the reported state-of-the-art on CQA by $10\%$.
For the sake of completeness, we also experimented with the reverse of this approach wherein the model first makes the predictions and then generates explanations based on those labels, which we call rationalization and is discussed below.

\paragraph{Rationalization}
In rationalization, the LM model conditions on the predicted labels along with the input to generate post-hoc rationalizations. So, during the fine-tuning step, the input context contains the output label and is constructed as follows:
\begin{center}
$C_{RA} =$ `` $q$, $c0$, $c1$, or $c2$? $a$ because "
\end{center}
The training objective for the LM in rationalization is similar to that in reasoning except that in this case, the model has access to the ground truth labels to the input questions during training.
Because the language model is conditioned on the predicted label, 
the explanations cannot be considered as common sense reasoning. Instead, they offer a {\it rationalization} that makes the model more accessible and interpretable. 
We find that this approach outperforms the current best model by $6\%$ and also produces interestingly good quality explanations as discussed in Section~\ref{experiments}.

For CAGE, we generate sequences of maximum length $20$, use a batch size of $36$, train for a maximum of $10$ epochs, selecting the best model based on validation BLEU and perplexity scores. Learning rate was set to $1e^{-6}$, warmed up linearly with proportion $0.002$ and weight decay $0.01$.

\subsection{Commonsense Predictions with Explanations}
\vspace{-0.2cm}
Given either a human explanation from CoS-E or reasoning from a language model, 
we can then learn to perform predictions on the CQA task.
For the classification module of our proposed approach,
we adopt the widely popular BERT model \citep{devlin2018bert} which we refer to as just BERT.
BERT can be fine-tuned for multiple choice question answering by adding a simple binary classifier that takes as input the final state corresponding to the the special \texttt{[CLS]} token placed at the start of all inputs to BERT models \citep{devlin2018bert}. 
We apply this same approach to the CQA task. 
For each example in the dataset, we construct three input sequences for fine-tuning BERT.
Each sequence is the concatenation of the question, a separator token \texttt{[SEP]}, and one of the answer choices.
If the approach requires explanation from either CoS-E or automatically generated as in the CAGE,
we concatenate the question, \texttt{[SEP]}, the explanation, \texttt{[SEP]}, and an answer choice. For BERT, the explanations share the same input representation as that of the questions. We also experimented with the explanation sharing the same representation as that of the answer choice but found that the performance decreased slightly.

When explanations are used only during training, 
the explanation variable is optional and the answer choices directly follow the question during evaluation.
For all our experiments we used a train batch size of $24$, test batch size of $12$, $10$ training epochs and maximum sequence length of $50$ for the baseline and $175$ for all experiments involving explanations.
The right part of Figure~\ref{model} gives an overview of the classification module of our proposed approach.

\subsection{Transfer to out-of-domain datasets}
\vspace{-0.2cm}
Transfer without fine-tuning to out-of-domain NLP datasets is known to exhibit poor performance. For example, for the comparatively easier natural langauge inference task with fixed labels, \citet{bowman2015large} show that the accuracy dropped by \(25\%\) when training on SNLI and evaluating on SICK-E \citep{marelli-etal-2014-sick}. We study transfer of natural language explanations from the CQA to SWAG \citep{zellers2018swag} and Story Cloze Test \citep{mostafazadeh-EtAl:2016:N16-1}. Both the datasets are multiple-choice like CQA and the authors publicize them as commonsense reasoning and inference tasks. 

We use the GPT language model fine-tuned on CQA train and dev sets to generate explanations on the SWAG train and val sets (with \(73546\) and \(20006\) instances respectively) and the Story Cloze Spring 2016 val and test sets (with \(1870\) instances each). We then train a BERT classifier using the input instances and generated explanations and evaluate on the SWAG and Story Cloze test sets. 
\section{Experimental Results}
\vspace{-0.2cm}
\label{experiments}
We present results on the CQA dataset using variations of our proposed Commonsense Auto-Generated Explanations (CAGE).
All our models are based on BERT, which also serves as our baseline without any CoS-E or CAGE.
All our ablation analysis is conducted on the CQA dev-random-split. 
We also show results for key models on the final test split.\footnote{\url{https://www.tau-nlp.org/csqa-leaderboard}} 

\begin{table}[!ht]
\centering
\small
\begin{tabular}{lc}
\toprule
\textbf{Method}&\textbf{Accuracy (\%)}\\
\midrule
BERT (baseline) & $63.8$\\
CoS-E-open-ended &$65.5$\\
CAGE-reasoning & $\mathbf{72.6}$ \\
\bottomrule
\end{tabular}
\caption{Results on CQA dev-random-split with CoS-E used during training.
}
\vspace{-0.5cm}
\label{table:results-dev-noexpl}
\end{table}

Table~\ref{table:results-dev-noexpl} shows results that compare a BERT baseline that uses only the CQA inputs and the same architecture but trained using inputs that contain explanations from CoS-E during training. 
The BERT baseline model reaches $64$\% accuracy and adding open-ended human explanations (CoS-E-open-ended) alongside the questions during training results in a $2\%$ boost in accuracy. 
By generating explanations as described in Section~\ref{sec:generating},
we can give the commonsense question answering model access to an explanation that is not conditioned on the ground truth.
These explanations (CAGE-reasoning) can be provided during both training and validation and increases the accuracy to $72\%$.


\begin{table}[!t]
\centering
\small
\begin{tabular}{lc}
\toprule
\textbf{Method}&\textbf{Accuracy (\%)}\\
\midrule
RC \citep{talmor2018commonsenseqa}& $47.7$\\
GPT \citep{talmor2018commonsenseqa} & $54.8$\\
CoS-E-open-ended &$60.2$\\
CAGE-reasoning  &$\mathbf{64.7}$\\
Human  \citep{talmor2018commonsenseqa}& $95.3$\\
\bottomrule
\end{tabular}
\caption{Test accuracy on CQA v1.0. The addition of CoS-E-open-ended during training dramatically improves performance. Replacing CoS-E during training with CAGE reasoning during both training and inference leads to an absolute gain of $10$\% over the previous state-of-the-art.}
\vspace{-0.25cm}
\label{table:results-test}
\end{table}

Table~\ref{table:results-test} shows the results obtained on the CQA test split. 
We report our two best models that represent using human explanations (CoS-E-open-ended) for training only and using language model explanations (CAGE-reasoning) during both train and test. 
We compare our approaches to the best reported models for the CQA task \citep{talmor2018commonsenseqa}. 
We observe that using CoS-E-open-ended during training improves the state-of-the-art by approximately $6\%$. 

\citet{talmor2018commonsenseqa} experimented with using Google search of ``question + answer choice"  for each example in the dataset and collected $100$ top snippets per answer choice to be used as context for their Reading Comprehension (RC) model. 
They found that providing such extra data does not improve accuracy. 
On the other hand, using CAGE-reasoning resulted in a gain of $10\%$ accuracy over the previous state-of-the-art.
This suggests that our CoS-E-open-ended and CAGE-reasoning explanations provide far more useful information than what can be achieved through simple heuristics like using Google search to find relevant snippets. We observed
that our models' performance on test is lower
than those on validation and this trend was confirmed by the organizers
of the task.

\begin{table}[!t]
\centering
\small
\begin{tabular}{lc}
\toprule
\textbf{Method}&\textbf{Accuracy (\%)}\\
\midrule
CoS-E-selected w/o ques &$53.0$\\
CoS-E-limited-open-ended &$67.6$\\
CoS-E-selected &$70.0$\\
CoS-E-open-ended w/o ques  & $84.5$\\
CoS-E-open-ended*  &$\mathbf{89.8}$\\
\bottomrule
\end{tabular}
\caption{Oracle results on CQA dev-random-split using different variants of CoS-E for both training and validation. * indicates CoS-E-open-ended used during both training and validation to contrast with CoS-E-open-ended used only during training in Table~\ref{table:results-dev-noexpl}.}
\vspace{-0.4cm}
\label{table:results-dev}
\end{table}

To establish an oracle upper-bound on the performance, we also explored an experimental setting in which human-generated explanations from CoS-E are provided during both training and validation. 
These results are summarized in Table~\ref{table:results-dev}.
We note that this is an unfair setting because the human that provided the explanation had access to the ground truth answer;
these results merely serve as an oracle for how much potential benefit can come from using CoS-E-open-ended.
If the open-ended human explanations (CoS-E-open-ended) are provided at inference time, performance jumps to approximately $90\%$.
 These results also motivate an attempt to automatically
generate explanations that establish the
world knowledge needed to solve CQA. CAGE-reasoning is our attempt towards this goal.


Table~\ref{table:results-dev} also contains results that use only the explanation and exclude the original question from CQA denoted by `w/o question'.
These variants also use explanation during both train and validation.
For these experiments we give the explanation in place of the question followed by the answer choices as input to the model. 
When the explanation consists of words humans selected as justification for the answer (CoS-E-selected), 
the model was able to obtain $53\%$ in contrast to the $85$\% achieved by the open-ended human explanations (CoS-E-open-ended).
Adding the question boosts performance for CoS-E-selected to $70$\%, again falling short of almost $90$\% achieved by CoS-E-open-ended. 
We conclude then that our full, open-ended CoS-E thus supply a significant source of information beyond simply directing the model towards the most useful information already in the question.

\begin{table}[!ht]
\centering
\small
\begin{tabular}{lc}
\toprule
\textbf{Method}&\textbf{Accuracy (\%)}\\
\midrule
CAGE-reasoning &$55.7$\\
BERT baseline & $56.7$\\
CoS-E-open-ended &$\mathbf{58.2}$\\
\bottomrule
\end{tabular}
\caption{Test results on CQA v1.11.}
\vspace{-0.4cm}
\label{table:results-v1.11}
\end{table}

We experimented with one final setting in which we only used open-ended explanations that did not contain {\it any} word from {\it any} answer choices ($23\%$. 
In this setting, we call these ``CoS-E-limited-open-ended" explanations because these explanations are limited in the choice of words allowed.
We observe that even using these limited kind of explanations improves over the BERT baseline in Table~\ref{table:results-dev},
which suggests that the explanations are providing useful information beyond just mentioning the correct or incorrect answers.

We also evaluated our key models -- CoS-E-open-ended used during training only and the CAGE reasoning on the v1.11 of CQA that was released before the final submission. Table~\ref{table:results-v1.11} shows the results obtained on the more challenging CQA v1.11. 

\citet{camburu2018snli} empirically show that transferring explanations on the natural language inference (NLI) problem from SNLI to MultiNLI performs very poorly and is still an open challenging problem. We study transfer of explanations on commonsense reasoning tasks. The NLI problem has a small fixed set of pre-defined labels unlike the commonsense reasoning tasks such as CQA, SWAG and Story Cloze. Table~\ref{table:results-transfer} shows the results obtained by the BERT baseline without explanations and using our transferred explanations from CQA to SWAG and Story Cloze. We observed that adding explanations led to a very small decrease ($<0.6\%$) in the performance compared to the baseline for both tasks.

\begin{table}[!ht]
\centering
\small
\begin{tabular}{lcc}
\toprule
\textbf{Method}&\textbf{SWAG}&\textbf{Story Cloze}\\
\midrule
BERT&84.2&89.8\\
\qquad + expl transfer&83.6&89.5\\
\bottomrule
\end{tabular}
\caption{Results for explanation transfer from CQA to out-of-domain SWAG and Sotry Cloze tasks.}
\vspace{-0.5cm}
\label{table:results-transfer}
\end{table}

\begin{table}[!t]
\centering
\scriptsize
\setlength{\tabcolsep}{1pt}
\begin{tabular}{ll}
\toprule
Question: &What could people do that involves talking?\\ 
Choices: &\textbf{confession}, carnival, state park \\
CoS-E: & confession is the only vocal action.\\
Reason & people talk to each other\\
Rationale: & people talk to people\\
\midrule
Question: &A child wants to play, what would they likely want?\\ 
Choices: &\textbf{play tag}, breathe, fall down\\
CoS-E: &A child to play tag \\
Reason &Children want to play tag, and they want to play tag with their\\
&friends.\\
Rationale: &Children want to play tag, what would they want to do?\\
\midrule
Question: &They were getting ready for a really long hike, he put the food \\
&in his what?\\ 
Choices: &recycling center, house, \textbf{backpack}\\
CoS-E: & Backpacks are used on hikes\\
Reason &a backpack is a place to store food and supplies.\\
Rationale: & a backpack is used to carry food and supplies\\
\midrule
Question: &You can do knitting to get the feeling of what?\\ 
Choices: &\textbf{relaxation}, your, arthritis \\
CoS-E: & Your are focusing on a repetitive task.\\
Reason &knitting is the only thing that is relaxing.\\
Rationale: &you can do knitting to get the feeling of what?\\
\bottomrule
\end{tabular}
\caption{Random sample of explanations generated by humans from CoS-E and our CAGE framework's reasoning and rationalization approaches. Boldface indicates gold label. All the typos and grammatical errors are as they appear in the actual output sequence.}
\vspace{-0.5cm}
\label{table:head_to_head}
\end{table}
\section{Analysis and Discussion}
\vspace{-0.2cm}
In Table~\ref{table:results-dev-noexpl}, using CAGE-reasoning at both train and validation resulted in an accuracy of $72\%$, 
but Table~\ref{table:results-dev} shows that if CAGE-reasoning truly captured all information provided in CoS-E-open-ended, performance would be $90\%$. 
This gap between CAGE and CoS-E prompted further analysis.

We measure quality of CAGE using human evaluation and automated metrics. 
One of the metrics is the BLEU score \citep{papineni2002bleu}, 
which measures syntactical precision by n-gram overlap. 
We also report perplexity, which provides a token-level measure of how well the language models predict the next word.
We obtained a peak BLEU score of $4.1$ between CAGE-reasoning and CoS-E-open-ended and perplexity of  $32$.
Language models that are not fine-tuned achieve BLEU score of only $0.8$.
Though it is clearly beneficial to fine-tune the LM and empirical results suggested that CAGE increased performance,
these scores suggest that humans and LMs have widely varying ways of providing useful explanations.

Error analysis on the baseline BERT model that does not use any explanations indicates that the model performs poorly on questions that are longer on an average and are more compositional. The average length of such questions is $14$ words as opposed to the average length of $13$ words for questions that the  model using CAGE predicts incorrectly. Therefore, we can conclude that explanations help elucidate the longer and more complicated compositional questions.

Table~\ref{table:head_to_head} shows a collection of examples from CQA, CoS-E, and CAGE samples.
We observe that CAGE-reasoning typically employs a much simpler construction than CoS-E-open-ended.
Nonetheless, this simple declarative mode can sometimes be more informative than CoS-E-open-ended.
CAGE achieves this by either providing more explicit guidance (as in the final example of Table~\ref{table:head_to_head})
or by adding meaningful context (as in the third example by introducing the word `friends').
We observe that CAGE-reasoning contains at least one of the answer choices $43\%$ of the time, 
out of which it contains the model's actual predicted answer choice $21\%$ of the time.
This suggests that there is more to the effectiveness of CAGE-reasoning than directly pointing to the answer. 

\begin{table}[!ht]
\centering
\setlength{\tabcolsep}{3pt}
\scriptsize
\begin{tabular}{ll}
\toprule
Question: &What is the main purpose of having a bath?\\
Choices:& \textbf{cleanness}, use water, exfoliation, \ul{hygiene}, wetness\\
Explanation: &the only purpose of having a bath is to clean yourself.\\
\midrule
Question: &Where can you store you spare linens near your socks?\\
Choices:& cabinet, \ul{chest}, hospital, \textbf{dresser drawers}, home\\
Explanation: &dresser drawer is the only place that you can store linens.\\
\midrule
Question: &Where do you find the most amount of leafs?,\\
Choices:& \textbf{forrest}, floral arrangement, \ul{compost pile}, field, ground\\
Explanation: &the most likely place to find leafs is in a garden.\\
\bottomrule
\end{tabular}
\caption{Random sample of incorrectly predicted instances by CAGE-reasoning on CQA v1.11 dev-set. Bold indicated ground-truth and underline indicates our CAGE's prediction.}
\vspace{-0.5cm}
\label{table:v1.11}
\end{table}

We also carried out human evaluations to compare $400$ examples of CoS-E and CAGE-reasoning. 
We asked human participants on Mechanical Turk to guess the most appropriate answer choice based on only the explanation without the question. 
This tests whether the explanation by itself is sufficient for a human to arrive at the same answer as the neural network. 
We found that Turkers were able to arrive at the same answer as the model based on CAGE-reasoning $42\%$ of the time.
This initially seemed low, but Turkers could only arrive at the same answer as humans using only CoS-E-open-ended $52$\% of the time

From Table~\ref{table:head_to_head}, we observed that CAGE-rationalization and CAGE-reasoning were often identical or differed only in word ordering or by replacing one of the answer choices with another.
Humans could predict the answer based on just CAGE-rationalization $42\%$ of the time, same as CAGE-reasoning. 
Although CAGE-rationalizations seem to be better than CAGE-reasoning, we
find that it does not drastically improve the model's
language generating behavior which is what humans
judge while trying to guess the right answer
without the actual question.

Even though CoS-E and CAGE are noisy, they empirically perform well when used by downstream models for CQA, but this is not the case for misleading explanations.
If we manually changed a random sample of $50$ examples to have adversarial misleading explanations, performance dropped from $60\%$ to $30\%$, well below the baseline of $50\%$ validation accuracy.
For example, we changed the explanation from ``being able to use`` to ``buying more will alleviate stress`` for the question ``If a couple is having financial issues, buying products can lead to what`` with answer choices ``economic boom", ``disagreements", ``being able to use".
Of the $70\%$ of the errors made by a model trained on misleading explanations, $57\%$ of them were instead correctly answered by our model trained with true CoS-E explanations.
This demonstrates the effectiveness of having well-informing explanations.

\begin{table*}[!ht]
\centering
\scriptsize
\begin{tabular}{ll}
\toprule
\multicolumn{2}{c}{\textbf{SWAG}}\\
\midrule
Question: &Men are standing on motorbikes getting ready for a motocross competition.\\
Choices:&man places the ladders onto a fence and winds up a marching wall, high with hammer and a stone., \textbf{man is talking to the camera and} \\
&\textbf{standing on a podium.}, man stands outside in the field going at arms of people and leading a long jumping calf in front., man drops  \\
&the javelin to the ground and jumps it very high.\\
Explanation: &man is talking to the camera and not the crowd.\\
\midrule
Question: &The man examines the instrument in his hand.\\
Choices:&The person studies a picture of the man playing the violin., \textbf{The person holds up the violin to his chin and gets ready.}, The person stops to\\
&speak to the camera again., The person puts his arm around the man and backs away.\\
Explanation: &the person is holding the instrument in his hand.\\
\midrule
Question: &The woman is seated facing the camera while another woman styles her hair.\\
Choices:&The woman in purple is wearing a blue dress and blue headband, using the pits to style her hair., The woman begins to cut the hair with her  \\
& hair then serves it and begins brushing her hair and styling it., The woman puts some right braids on his., \textbf{The woman continues to have }\\
&\textbf{her hair styled while turned away from the camera.}\\
Explanation: &the woman is using the braids to trim her hair.\\
\toprule
\multicolumn{2}{c}{\textbf{Story Cloze (ROCStories)}}\\
\midrule
Question: &My friends all love to go to the club to dance. They think it's a lot of fun and always invite. I finally decided to tag \\ 
& along last Saturday. I danced terribly and broke a friend's toe.\\
Choices:&My friends decided to keep inviting me out as I am so much fun., \textbf{The next weekend, I was asked to please stay home.}\\
Explanation: &the next weekend, i would be asked to stay home\\
\midrule
Question: &Ari spends $\$20$ a day on pickles. He decides to make his own to save money. He puts the pickles in brine. Ari waits 2 weeks for his pickles \\
& to get sour.\\
Choices:&\textbf{Ari opens the jar to find perfect pickles.}, Ari's pickles are sweet.\\
Explanation: &pickles are the only thing that can be found in a jar.\\
\midrule
Question: &Gina sat on her grandpa's bed staring outside. It was winter and his garden was dead until spring. Her grandpa had passed away so there \\
& would be no one to tend it. The weeds would take over and strangle the flowers.\\
Choices:&Gina asked her grandpa what kind of flowers he liked best., \textbf{Gina decided to go outside and pick some of the weeds.}\\
Explanation: &the weeds would take over and strangle the flowers.\\
\bottomrule
\end{tabular}
\caption{Random sample of explanations generated by the language model fine-tuned on CQA and transferred without further training to SWAG and Story Cloze. Bold indicates ground-truth.}
\vspace{-0.5cm}
\label{table:swag}
\end{table*}

\citet{camburu2018snli} use human explanations to train a neural network model on the SNLI dataset \citep{bowman2015large}. However, they obtain explanations at the cost of accuracy. The authors use the InferSent \citep{conneau2017supervised} model for classification and add a one-layer LSTM as the explanation decoder. They report a slight drop in performance ($< 1\%$) when training on human explanations and testing by first predicting an answer and then generating explanations. 
There is a further drop of approximately $2\%$ accuracy when their model generates explanations prior to predicting an answer based only on that explanations.
However, they also show that a bidirectional encoder with MLP-classifier obtains $96.83\%$ accuracy when given only human explanations. 
CQA experiences a lift from explanations when e-SNLI performance appears to degrade with explanations. For CQA, humans are able to predict the right answer only about $52\%$ of the time using only human explanations from CoS-E. 

On the more challenging CQA v1.11, we observed that our CoS-E model trained on human explanations but evaluated without explanations obtains state-of-the-art performance, beating the BERT baseline by \(1.5\%\). Surprisingly, we found that our CAGE-reasoning model performs slightly worse than the baseline. However, during error analysis we found that the language model explanations do not exhibit any obvious problems. Table~\ref{table:v1.11} shows some samples that CAGE predicts incorrectly. We observed that many of the incorrectly predicted instances had the correct answer in the generated explanation, such as ``dresser drawer" and ``cleanness" in the first two examples, but this information is not properly used by the BERT classifier. A more explicit method of guiding attention towards the relevant information in the explanations might be necessary for such cases. The model also frequently errs when the choices seem semantically close such as ``forest" and ``compost pile" in the third example. In these cases, the classifier often predicts the incorrect choice on v1.11, but was able to predict the correct choice on v1.0 when only $3$ choices were presented. This suggests that simply concatenating explanations is unable to make sufficiently clear the more difficult cases of the newer version of CQA.

Transferring the language model used to generate commonsense explanations to out-of-domain datasets, SWAG and Story Cloze, led to slight decrease in performance. Upon inspection, the generated explanations exhibited little grammatical or syntactical errors and often contained apparently relevant information. Table~\ref{table:swag} shows examples from both datasets and the corresponding generated explanations. In the SWAG dataset, each question is a video caption from activity recognition videos with choices about what might happen next and the correct answer is the video caption of the next scene. Generated explanations for SWAG appear to be grounded in the given images even though the language model was not at all trained on SWAG. Similarly, we found that for the Story Cloze dataset, the explanations had information pointing to the correct ending. Nonetheless, the classifier was unable to make use of this information to improve performance. 
\section{Conclusion and Future Work}
\vspace{-0.2cm}
We introduced the Common Sense Explanations
(CoS-E) dataset built on top of the existing CommonsenseQA
dataset. We also proposed
the novel Commonsense Auto-Generated Explanations (CAGE)
framework that trains a language model to
generate useful explanations when fine-tuned on
the problem input and human explanations 
These
explanations can then be used by a classifier model
to make predictions. We empirically show that
such an approach not only results in state-of-the-art
performance on a difficult commonsense reasoning task, but also opens further avenues for studying explanation as it relates to interpretable commonsense reasoning. We also performed comprehensive error analyses of language model explanations and evaluated explanation transfer to out-of-domain datasets. 

While CAGE focuses on generating explanations prior to predicting an answer, language models for explanation might also be jointly trained to predict the answer. They might also be extended to a broader set of tasks. With a sufficient dataset of explanations (analogous to CoS-E) for many tasks, it might be possible to fine-tune a more general explanatory language model that generates more useful explanations for unseen tasks.

With deferral of explanation to neural models, it will be crucial in the future to study the ethical implications of biases that are accumulated during pretraining or fine-tuning. Explanations must be carefully monitored to ensure that they do not reinforce negative or otherwise harmful reasoning that might then propagate into downstream models. For example, in CQA we observed significant gender disparity and bias with higher proportion of female pronouns used in negative contexts. This kind of bias has inevitably propagated into CoS-E and advise these datasets and trained models be used with that in mind.
\section*{Acknowledgements}
We would like to thank Melvin Gruesbeck for the illustration of CAGE in Figure 1. We also thank the anonymous reviewers for their feedback.
\bibliography{acl2019}

\begin{thebibliography}{29}
\expandafter\ifx\csname natexlab\endcsname\relax\def\natexlab#1{#1}\fi

\bibitem[{Antol et~al.(2015)Antol, Agrawal, Lu, Mitchell, Batra, Zitnick, and
  Parikh}]{vqa}
Stanislaw Antol, Aishwarya Agrawal, Jiasen Lu, Margaret Mitchell, Dhruv Batra,
  C.~Lawrence Zitnick, and Devi Parikh. 2015.
\newblock {VQA}: {V}isual {Q}uestion {A}nswering.
\newblock In \emph{International Conference on Computer Vision (ICCV)}.

\bibitem[{Bowman et~al.(2015)Bowman, Angeli, Potts, and
  Manning}]{bowman2015large}
Samuel~R Bowman, Gabor Angeli, Christopher Potts, and Christopher~D Manning.
  2015.
\newblock A large annotated corpus for learning natural language inference.
\newblock In \emph{Proceedings of the 2015 Conference on Empirical Methods in
  Natural Language Processing (EMNLP2015)}, pages 632--642.

\bibitem[{Camburu et~al.(2018)Camburu, Rockt{\"a}schel, Lukasiewicz, and
  Blunsom}]{camburu2018snli}
Oana-Maria Camburu, Tim Rockt{\"a}schel, Thomas Lukasiewicz, and Phil Blunsom.
  2018.
\newblock {e-SNLI: Natural Language Inference with Natural Language
  Explanations}.
\newblock In \emph{Advances in Neural Information Processing Systems
  (NeurIPS2018)}, pages 9560--9572.

\bibitem[{Conneau et~al.(2017)Conneau, Kiela, Schwenk, Barrault, and
  Bordes}]{conneau2017supervised}
Alexis Conneau, Douwe Kiela, Holger Schwenk, Lo{\"\i}c Barrault, and Antoine
  Bordes. 2017.
\newblock Supervised learning of universal sentence representations from
  natural language inference data.
\newblock In \emph{Proceedings of the 2017 Conference on Empirical Methods in
  Natural Language Processing (EMNLP2017)}, pages 670--680.

\bibitem[{Dai and Le(2015)}]{dai2015semi}
Andrew~M Dai and Quoc~V Le. 2015.
\newblock Semi-supervised sequence learning.
\newblock In \emph{Proceedings of the 28th International Conference on Neural
  Information Processing Systems (NIPS2015)}, pages 3079--3087. MIT Press.

\bibitem[{Devlin et~al.(2019)Devlin, Chang, Lee, and
  Toutanova}]{devlin2018bert}
Jacob Devlin, Ming-Wei Chang, Kenton Lee, and Kristina Toutanova. 2019.
\newblock \href {https://www.aclweb.org/anthology/N19-1423} {{BERT}:
  Pre-training of deep bidirectional transformers for language understanding}.
\newblock In \emph{Proceedings of the 2019 Conference of the North {A}merican
  Chapter of the Association for Computational Linguistics: Human Language
  Technologies, Volume 1 (Long and Short Papers)}, pages 4171--4186,
  Minneapolis, Minnesota. Association for Computational Linguistics.

\bibitem[{Hancock et~al.(2018)Hancock, Varma, Wang, Bringmann, Liang, and
  R{\'e}}]{hancock2018training}
Braden Hancock, Paroma Varma, Stephanie Wang, Martin Bringmann, Percy Liang,
  and Christopher R{\'e}. 2018.
\newblock Training classifiers with natural language explanations.
\newblock In \emph{Proceedings of the 56th Annual Meeting of the Association
  for Computational Linguistics (ACL2018)}, volume~1, pages 1884--1895.

\bibitem[{Howard and Ruder(2018)}]{howard2018universal}
Jeremy Howard and Sebastian Ruder. 2018.
\newblock Universal language model fine-tuning for text classification.
\newblock In \emph{Proceedings of the 56th Annual Meeting of the Association
  for Computational Linguistics (ACL2018)}, pages 328--339.

\bibitem[{Lei et~al.(2016)Lei, Barzilay, and Jaakkola}]{lei2016rationalizing}
Tao Lei, Regina Barzilay, and Tommi Jaakkola. 2016.
\newblock Rationalizing neural predictions.
\newblock In \emph{Proceedings of the 2016 Conference on Empirical Methods in
  Natural Language Processing (EMNLP2016)}, pages 107--117.

\bibitem[{Levesque et~al.(2012)Levesque, Davis, and
  Morgenstern}]{levesque2012winograd}
Hector Levesque, Ernest Davis, and Leora Morgenstern. 2012.
\newblock The winograd schema challenge.
\newblock In \emph{Thirteenth International Conference on the Principles of
  Knowledge Representation and Reasoning}.

\bibitem[{Marelli et~al.(2014)Marelli, Menini, Baroni, Bentivogli, Bernardi,
  and Zamparelli}]{marelli-etal-2014-sick}
Marco Marelli, Stefano Menini, Marco Baroni, Luisa Bentivogli, Raffaella
  Bernardi, and Roberto Zamparelli. 2014.
\newblock \href
  {http://www.lrec-conf.org/proceedings/lrec2014/pdf/363_Paper.pdf} {A {SICK}
  cure for the evaluation of compositional distributional semantic models}.
\newblock In \emph{Proceedings of the Ninth International Conference on
  Language Resources and Evaluation ({LREC}{'}14)}, pages 216--223, Reykjavik,
  Iceland. European Language Resources Association (ELRA).

\bibitem[{McCann et~al.(2017)McCann, Bradbury, Xiong, and
  Socher}]{mccann2017learned}
Bryan McCann, James Bradbury, Caiming Xiong, and Richard Socher. 2017.
\newblock Learned in translation: Contextualized word vectors.
\newblock In \emph{Advances in Neural Information Processing Systems}, pages
  6294--6305.

\bibitem[{McCann et~al.(2018)McCann, Keskar, Xiong, and
  Socher}]{mccann2018natural}
Bryan McCann, Nitish~Shirish Keskar, Caiming Xiong, and Richard Socher. 2018.
\newblock The natural language decathlon: Multitask learning as question
  answering.
\newblock \emph{arXiv preprint arXiv:1806.08730}.

\bibitem[{Mikolov et~al.(2013)Mikolov, Chen, Corrado, and
  Dean}]{mikolov2013efficient}
Tomas Mikolov, Kai Chen, Greg Corrado, and Jeffrey Dean. 2013.
\newblock Efficient estimation of word representations in vector space.
\newblock \emph{arXiv preprint arXiv:1301.3781}.

\bibitem[{Mostafazadeh et~al.(2016)Mostafazadeh, Chambers, He, Parikh, Batra,
  Vanderwende, Kohli, and Allen}]{mostafazadeh-EtAl:2016:N16-1}
Nasrin Mostafazadeh, Nathanael Chambers, Xiaodong He, Devi Parikh, Dhruv Batra,
  Lucy Vanderwende, Pushmeet Kohli, and James Allen. 2016.
\newblock \href {http://www.aclweb.org/anthology/N16-1098} {A corpus and cloze
  evaluation for deeper understanding of commonsense stories}.
\newblock In \emph{Proceedings of the 2016 Conference of the North American
  Chapter of the Association for Computational Linguistics: Human Language
  Technologies (NAACL2016)}, pages 839--849, San Diego, California. Association
  for Computational Linguistics.

\bibitem[{Papineni et~al.(2002)Papineni, Roukos, Ward, and
  Zhu}]{papineni2002bleu}
Kishore Papineni, Salim Roukos, Todd Ward, and Wei-Jing Zhu. 2002.
\newblock Bleu: a method for automatic evaluation of machine translation.
\newblock In \emph{Proceedings of the 40th Annual meeting on Association for
  Computational Linguistics (ACL2002)}, pages 311--318. Association for
  Computational Linguistics.

\bibitem[{Pennington et~al.(2014)Pennington, Socher, and
  Manning}]{pennington2014glove}
Jeffrey Pennington, Richard Socher, and Christopher Manning. 2014.
\newblock Glove: Global vectors for word representation.
\newblock In \emph{Proceedings of the 2014 conference on empirical methods in
  natural language processing (EMNLP2014)}, pages 1532--1543.

\bibitem[{Peters et~al.(2018)Peters, Neumann, Iyyer, Gardner, Clark, Lee, and
  Zettlemoyer}]{peters2018deep}
Matthew~E Peters, Mark Neumann, Mohit Iyyer, Matt Gardner, Christopher Clark,
  Kenton Lee, and Luke Zettlemoyer. 2018.
\newblock Deep contextualized word representations.
\newblock \emph{arXiv preprint arXiv:1802.05365}.

\bibitem[{Radford et~al.(2018)Radford, Narasimhan, Salimans, and
  Sutskever}]{radford2018improving}
Alec Radford, Karthik Narasimhan, Tim Salimans, and Ilya Sutskever. 2018.
\newblock Improving language understanding by generative pre-training.
\newblock
  \emph{\text{https://s3-us-west-2.amazonaws.com/openai-assets/}\\\text{research-covers/language-unsupervised/}\\\text{language\_understanding\_paper.pdf}}.

\bibitem[{Rajani and Mooney(2018)}]{rajani2018stacking}
Nazneen~Fatema Rajani and Raymond Mooney. 2018.
\newblock Stacking with auxiliary features for visual question answering.
\newblock In \emph{Proceedings of the 2018 Conference of the North American
  Chapter of the Association for Computational Linguistics: Human Language
  Technologies, Volume 1 (Long Papers)}, volume~1, pages 2217--2226.

\bibitem[{Rajani and Mooney(2017)}]{rajani:vigil17}
Nazneen~Fatema Rajani and Raymond~J. Mooney. 2017.
\newblock \href
  {http://www.cs.utexas.edu/users/ai-lab/pub-view.php?PubID=127684} {Ensembling
  visual explanations for vqa}.
\newblock In \emph{Proceedings of the NIPS 2017 workshop on Visually-Grounded
  Interaction and Language (ViGIL)}.

\bibitem[{Seo et~al.(2017)Seo, Kembhavi, Farhadi, and
  Hajishirzi}]{Seo2017BidirectionalAF}
Min~Joon Seo, Aniruddha Kembhavi, Ali Farhadi, and Hannaneh Hajishirzi. 2017.
\newblock Bidirectional attention flow for machine comprehension.
\newblock \emph{CoRR}, abs/1611.01603.

\bibitem[{Talmor et~al.(2019)Talmor, Herzig, Lourie, and
  Berant}]{talmor2018commonsenseqa}
Alon Talmor, Jonathan Herzig, Nicholas Lourie, and Jonathan Berant. 2019.
\newblock \href {https://www.aclweb.org/anthology/N19-1421} {{C}ommonsense{QA}:
  A question answering challenge targeting commonsense knowledge}.
\newblock In \emph{Proceedings of the 2019 Conference of the North {A}merican
  Chapter of the Association for Computational Linguistics: Human Language
  Technologies, Volume 1 (Long and Short Papers)}, pages 4149--4158,
  Minneapolis, Minnesota. Association for Computational Linguistics.

\bibitem[{Trinh and Le(2018)}]{trinh2018simple}
Trieu~H Trinh and Quoc~V Le. 2018.
\newblock A simple method for commonsense reasoning.
\newblock \emph{arXiv preprint arXiv:1806.02847}.

\bibitem[{Vaswani et~al.(2017)Vaswani, Shazeer, Parmar, Uszkoreit, Jones,
  Gomez, Kaiser, and Polosukhin}]{vaswani2017attention}
Ashish Vaswani, Noam Shazeer, Niki Parmar, Jakob Uszkoreit, Llion Jones,
  Aidan~N Gomez, {\L}ukasz Kaiser, and Illia Polosukhin. 2017.
\newblock Attention is all you need.
\newblock In \emph{Advances in Neural Information Processing Systems
  (NIPS2017)}, pages 5998--6008.

\bibitem[{Wang et~al.(2018)Wang, Singh, Michael, Hill, Levy, and
  Bowman}]{wang2018glue}
Alex Wang, Amapreet Singh, Julian Michael, Felix Hill, Omer Levy, and Samuel~R
  Bowman. 2018.
\newblock Glue: A multi-task benchmark and analysis platform for natural
  language understanding.
\newblock \emph{arXiv preprint arXiv:1804.07461}.

\bibitem[{Winograd(1972)}]{winograd1972understanding}
Terry Winograd. 1972.
\newblock Understanding natural language.
\newblock \emph{Cognitive psychology}, 3(1):1--191.

\bibitem[{Zellers et~al.(2018)Zellers, Bisk, Schwartz, and
  Choi}]{zellers2018swag}
Rowan Zellers, Yonatan Bisk, Roy Schwartz, and Yejin Choi. 2018.
\newblock Swag: A large-scale adversarial dataset for grounded commonsense
  inference.
\newblock In \emph{Proceedings of the 2018 Conference on Empirical Methods in
  Natural Language Processing (EMNLP2018)}, pages 93--104.

\bibitem[{Zhong et~al.(2018)Zhong, Tang, Duan, Zhou, Wang, and
  Yin}]{zhong2018improving}
Wanjun Zhong, Duyu Tang, Nan Duan, Ming Zhou, Jiahai Wang, and Jian Yin. 2018.
\newblock Improving question answering by commonsense-based pre-training.
\newblock \emph{arXiv preprint arXiv:1809.03568}.

\end{thebibliography}
\bibliographystyle{acl_natbib}

\end{document}